\documentclass{article}
\usepackage{spconf,amsmath,graphicx}
\usepackage{times}
\usepackage{url}
\usepackage{latexsym}
\usepackage{multirow}
\usepackage{amssymb, bm}
\usepackage{arydshln}
\usepackage{color}

\usepackage{CJK}

\newcommand{\SmpNluDataset}{ECDT-NLU}
\newcommand{\seqLevel}{sentence-level}
\newcommand{\charLevel}{character-level}

\newcommand{\newcite}[2]{#1 et al.~\cite{#2}}
\newcommand{\parjump}{\vspace{+0.2em}}
\newcommand{\newparagraph}[1]{\par \parjump \noindent \textbf{#1}}

\title{Injecting Word Information with Multi-Level Word Adapter for Chinese Spoken Language Understanding}
\name{Dechuan Teng$^{\star}$ \qquad Libo Qin$^{\star}$ \qquad Wanxiang Che$^{\dagger}$ \qquad Sendong Zhao \qquad Ting Liu
\thanks{$^{\star}$ Equal contributions.}
\thanks{$^{\dagger}$ Corresponding author.}}
\address{Research Center for Social Computing and Information Retrieval, Harbin Institute of Technology, China}
\begin{document}
%
\maketitle
\begin{CJK*}{UTF8}{gbsn}

\begin{abstract}
In this paper, we improve Chinese spoken language understanding (SLU) by injecting word information.
Previous studies on Chinese SLU do not consider the word information, failing to detect word boundaries that are beneficial for intent detection and slot filling.
To address this issue, we propose a multi-level word adapter to inject word information for Chinese SLU, which consists of
(1) \textit{\seqLevel{}} word adapter, which directly fuses the sentence representations of the word information and character information to perform intent detection and
(2) \textit{\charLevel{}} word adapter, which is applied at each character for selectively controlling weights on word information as well as character information.
Experimental results on two Chinese SLU datasets show that our model can capture useful word information and achieve state-of-the-art performance.
\end{abstract}

\begin{keywords}
Chinese Spoken Language Understanding, Intent Detection, Slot Filling, Word Adapter
\end{keywords}
\section{Introduction}
\label{sec:intro}

Spoken language understanding plays an important role in task-oriented dialog systems, which mainly consists of two subtasks including slot filling and intent detection~\cite{tur2011spoken}.
Take the sentence ``\textit{use netflix to play music}'' as an example, intent detection aims to classify the intent label (i.e., \texttt{PlayMusic}) for the whole sentence, while slot filling aims to assign different slot labels (i.e., \texttt{O}, \texttt{B-service}, \texttt{O}, \texttt{O}, \texttt{O}) for each token in this sentence.

Compared with SLU in English, Chinese SLU faces a unique challenge since it usually needs word segmentation.
Nevertheless, the imperfect segmentation performed by the CWS (Chinese Word Segmentation) system will misidentify slot boundaries and predict wrong slot categories, therefore suffers from the error propagation.
To address this issue,~\newcite{Liu}{liu2019cm} proposed a character-based method to perform Chinese SLU in a joint model at the character level, achieving state-of-the-art performance. 

However, the main drawback of the character-based SLU model is that explicit word sequence information is not fully exploited, which might be useful for understanding Chinese texts.
Take the Chinese utterance ``周冬雨~(Dongyu Zhou) / 有~(has) / 哪些~(which) / 电影~(movies)'' as an example, where the utterance is split into a sequence of words by ``/''.
The character-based model is easily confused and predicts ``周~(week)'' as \texttt{Datetime\_date} and treats ``冬雨~(winter rain)'' as \texttt{Datetime\_time} wrongly.
In contrast, with the help of words such as ``周冬雨~(Dongyu Zhou)'', a model can easily detect the correct slot label \texttt{Artist}.
Thus, it's important to incorporate words into Chinese SLU tasks.
Unfortunately, there has been a limited amount of research on how to introduce word information to enhance Chinese SLU effectively.

To bridge the gap, we propose a simple but effective method for injecting word information into Chinese SLU.
Since SLU is unique in jointly modeling two correlated tasks, exploiting their interaction can be useful for modeling fine-grained word information transfer between two tasks.
To this end, we design a multi-level word adapter to perform slot filling and intent detection in a joint model: (1) \textit{\seqLevel{}} word adapter directly fuses the word-aware and character-aware sentence representations for identifying intents;
(2) \textit{\charLevel{}} word adapter adaptively determines the weights between character features and word features for assigning slot labels for each character, so that fine-grained combinations of word knowledge can be achieved.
In addition, the proposed word adapter is applied as a plugin for the output layer, with no need to change other components in the original character-based model, achieving more flexibility.

The experimental results show the effectiveness of our framework by outperforming the SOTA methods by a large margin.
All datasets and codes will be publicly available at \url{https://github.com/AaronTengDeChuan/MLWA-Chinese-SLU}.
\section{Approach}
\label{sec:approach}

\begin{figure*} [ht]
	\centering
	\includegraphics[scale=0.45]{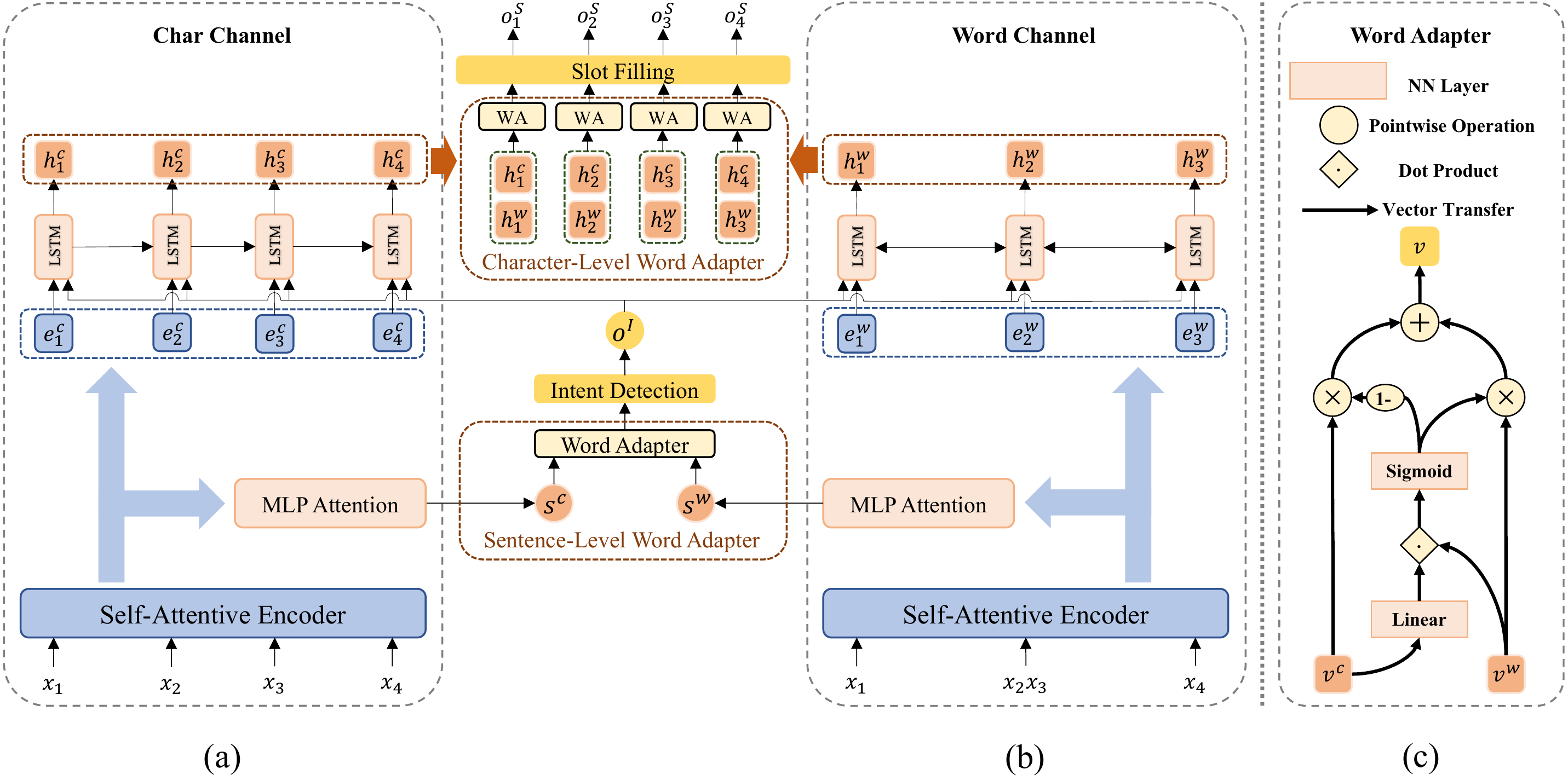}	
	\caption{Illustration of our proposed Multi-Level Word Adapter for Chinese SLU, which consists of a \seqLevel{} word adapter for intent detection and a \charLevel{} word adapter for slot filling. The internal structure of a word adapter is shown in (c).}
	\label{fig:framework}
\end{figure*}

\subsection{Overview}
We build our model based on a vanilla character-based model, as shown in Figure~\ref{fig:framework} (a).
To explicitly incorporate word information, as shown in \ref{fig:framework} (b), we propose a multi-level word adapter to capture \textit{sentence-level} and \textit{character-level} word information for intent detection and slot filling.
\subsection{Vanilla Character-based Model}
The vanilla character-based model, as shown in Figure~\ref{fig:framework}(a), consists of a shared self-attentive encoder, an intent detection decoder, and a slot filling decoder.

\newparagraph{Char-Channel Encoder}
Following~\newcite{Qin}{qin2019stack}, we adopt a shared self-attentive encoder to obtain the character encoding representations.
It mainly consists of a self-attention mechanism to extract the contextual information and a BiLSTM~\cite{hochreiter1997long} to capture sequential information within characters.

Formally, the BiLSTM takes the input utterance as input, conduct recurrent steps forwardly and backwardly, and obtain context-sensitive sequential features.
The idea of the self-attention module from~\newcite{Vaswani}{vaswani2017attention} is to capture contextualized representation for each token in a sequence that has shown effective in SLU tasks~\cite{qin2019stack}.
Finally, given a Chinese utterance $ \mathbf{c}=\left\{ c_1,c_2,\cdots,c_N \right\} $ with $ N $ characters, we concatenate the output of BiLSTM and self-attention over character sequence $ \mathbf{c} $ to obtain the final encoding representations $ \mathbf{E}^c $ = $\left\{ \mathbf{e}^c_1,\mathbf{e}^c_2,\cdots,\mathbf{e}^c_N \right\}$.

\newparagraph{Intent Detection}
We first apply a MLP attention module \cite{zhong2018global,zhang2018neural} to obtain the whole utterance representation $\mathbf{s}^c $ by calculating the weighted sum of all hidden states $ \mathbf{e}_t^c $ in $ \mathbf{E}^c $.
Then, we take the overall vector $ \mathbf{s}^c $ to perform intent detection as follows:
\begin{equation}
	\begin{aligned}
		P\left( \widetilde{y}=j \vert \mathbf{s}^c \right) &= \operatorname{softmax} \left( \mathbf{W}^{I} \mathbf{s}^c + \mathbf{b}^{I} \right)
		\\
		\mathbf{o}^{I} &= \mathop{\arg\max}_{\widetilde{y}\in S^{int}} P\left( \widetilde{y} \vert \mathbf{s}^c \right)
	\end{aligned}
	\label{eq:intent detection}
\end{equation}
where $ S^{int} $ is the intent label set, and $ \mathbf{W}^{I} $, $ \mathbf{b}^{I} $ are trainable parameters.

\newparagraph{Slot Filling}
We use a unidirectional LSTM as the slot filling decoder and follow prior works \cite{qin2019stack} to leverage the intent information to guide the slot prediction.
Previous studies have demonstrated the superiority of exploiting explicit information from closely correlated task~\cite{qin2019stack,zhao2019neural}. 
At each decoding step $ t $, the decoder hidden state $ \mathbf{h}^{c,S}_t $ is calculated as follows:
\begin{equation}
	\mathbf{h}^{c,S}_t = \mathbf{LSTM} \left( \mathbf{e}^c_t \oplus \phi ^{\text{int}} \left( \mathbf{o}^{I} \right) \oplus \mathbf{y}^{S}_{t-1},\mathbf{h}^{c,S}_{t-1} \right)
	\label{eq:slot filling decoder}
\end{equation}
where $ \phi ^{\text{int}} \left( \cdot \right) $ represents the embedding matrix of intents, and $ \mathbf{y}^{S}_{t-1} $ is the embedding of the emitted slot label from previous decoding step.
Then, the hidden state $ \mathbf{h}^{c,S}_t $ is utilized to perform slot filling:
\begin{equation}
	\begin{aligned}
		P\left( \widetilde{y}=j \vert \mathbf{h}^{c,S}_t \right) &= \operatorname{softmax} \left( \mathbf{W}^{S} \mathbf{h}^{c,S}_t + \mathbf{b}^{S} \right) \\
		\mathbf{o}^{S}_t &= \mathop{\arg\max}_{\widetilde{y}\in S^{slot}} P\left( \widetilde{y} \vert \mathbf{h}^{c,S}_t \right) \\
		\mathbf{y}^{S}_t &= \phi ^{\text{slot}} \left( \mathbf{o}^{S}_t \right)
	\end{aligned}
	\label{eq:slot filling}
\end{equation}
where $ \mathbf{W}^{S} $ and $ \mathbf{b}^{S} $ are trainable parameters, $ S^{slot} $ is the slot label set, $ \mathbf{o}^{S}_t $ is the slot label of $ t $th character $ c_t $, and $ \phi ^{\text{slot}} \left( \cdot \right) $ represents the embedding matrix of slots.

\subsection{Multi-Level Word Adapter}
The multi-level word adapter is the core module for combine character and word features, which can be used as a plugin to the above character-based model.
In particular, it consists of a word-channel encoder to obtain the word encoding information, a \textit{\seqLevel{}} word adapter to enhance intent detection at the sentence level, and a \textit{\charLevel{}} word adapter to boost slot filling at the character level.

\newparagraph{Word-Channel Encoder}
In our framework, the word-channel encoder is independent of the char-channel encoder, that is, we can freely decide how to utilize word information, without considering the interference between two encoders.

When using an external CWS system, we conduct word segmentation over the utterance $ \mathbf{c} $ and obtain the word sequences $ \mathbf{w}=\left\{ w_1,w_2,\cdots,w_M \right\} \left(M \leq N \right) $.
Same with char channel encoder, the word channel encoder generates the final word-channel representations $ \mathbf{E}^w $ = $\left\{ \mathbf{e}^w_1,\mathbf{e}^w_2,\cdots,\mathbf{e}^w_M \right\}$.

\newparagraph{Word Adapter}
The word adapter shown in Figure~\ref{fig:framework} (c) is a simple neural network layer for adaptively integrating character and word features.
Given the input character vector $ \mathbf{v}^c \in\mathbb{R}^{d} $ and word vector $ \mathbf{v}^w \in\mathbb{R}^{d} $, we calculate the weight between two input vectors and then conduct a weighted sum of these two vectors:
\begin{equation}
	\begin{aligned}
		\operatorname{WA} \left( \mathbf{v}^c, \mathbf{v}^w \right) &= \left( 1 - \lambda \right) \cdot \mathbf{v}^c + \lambda \cdot \mathbf{v}^w
		\\
		\lambda &= \operatorname{sigmoid}\left( {\mathbf{v}^c}^\top \mathbf{W}_{f} \mathbf{v}^w \right)
		\label{eq:word adapter}
	\end{aligned}
\end{equation}
where $ \mathbf{W}_{f} \in \mathbb{R}^{d\times d} $ is trainable parameters and $ \lambda $ can adaptively adjust the importance between character and word information.

\newparagraph{Sentence-Level Word Adapter}
Given contextualized representations $ \mathbf{E}^c $ and $ \mathbf{E}^w $ over character and word sequences, we first obtain two summary vectors $ \mathbf{s}^c $ and $ \mathbf{s}^w $ using the MLP attention module. 
Then, a \textit{\seqLevel{}} word adapter calculate the fused summarized vector $ \mathbf{v}^I = \operatorname{WA} \left( \mathbf{s}^c, \mathbf{s}^w \right) $ and then use it to predict the intent label $ \mathbf{o}^{I} $ with the Equation (\ref{eq:intent detection}).
\begin{equation}
	\begin{aligned}
		P\left( \widetilde{y}=j \vert \mathbf{v}^I \right) &= \operatorname{softmax} \left( \mathbf{W}^{I} \mathbf{v}^I + \mathbf{b}^{I} \right)
	\end{aligned}
	\label{eq:intent detection_2}
\end{equation}

\newparagraph{Character-Level Word Adapter}
%
For slot filling, we first adopt a bidirectional LSTM to strengthen slot-aware word representations.
At each time step $ i $, the hidden state $ \mathbf{h}^{w,S}_i = {\mathbf{BiLSTM}} \left( \mathbf{e}^w_i \oplus \phi ^{\text{int}} \left( \mathbf{o}^{I} \right),{\bf{h}}^{w,S}_{i-1} \right)$ is derived from the corresponding word representation $ \mathbf{e}^w_i $ and the embedding $ \phi ^{\text{int}} \left( \mathbf{o}^{I} \right) $ of the predicted intent $ \mathbf{o}^{I} $.

Then, a \textit{\charLevel{}} word adapter is adopted to determine different integration weights for different combinations of character and word features for each character:\footnote{
	$ len \left( \text{周冬雨} \right) = 3$,~~$ \mathbb{I}\left( True \right) = 1 $, and $ \mathbb{I}\left( False \right) = 0 $.
	Given a word sequence $ \mathbf{w}=\left\{ \text{``周冬雨''}, \text{``有''}, \text{``哪些''}, \text{``电影''} \right\} $, $ f_{align}(t, \mathbf{w}) $ gives the position index of the word corresponding to the $t$th character in $ \mathbf{w} $ (e.g., $ f_{align}(3, \mathbf{w}) = 1 $, $ f_{align}(4, \mathbf{w}) = 2 $, $ f_{align}(6, \mathbf{w}) = 3 $).
}
\begin{equation}
	\begin{aligned}
		\mathbf{v}^{S}_t = \operatorname{WA} \left( \mathbf{h}^{c,S}_t, \mathbf{h}^{w,S}_{f_{align}(t, \mathbf{w})} \right)
	\end{aligned}
\end{equation}

Finally, we utilize the integration representation $ \mathbf{v}^{S}_t $ that contains the word information and character information instead of $\mathbf{h}_t^{c,S}$ to perform slot filling with the Equation~(\ref{eq:slot filling}):
\begin{equation}
	\begin{aligned}
		P\left( \widetilde{y}=j \vert \mathbf{v}^{S}_t \right) = \operatorname{softmax} \left( \mathbf{W}^{S} \mathbf{v}^{S}_t + \mathbf{b}^{S} \right)
	\end{aligned}
	\label{eq:slot filling_2}
\end{equation}

\newparagraph{Joint Training}
Following~\newcite{Goo}{goo2018slot}, we adopt a joint training scheme for optimization, where the final joint objective function is computed as follows:
\begin{equation}
	\begin{aligned}
		& \mathcal{L} = - \log P\left( \hat{y}^{I} \vert \mathbf{v}^{I} \right) 
		- 
		\sum_{i=1}^{N} \log P\left( \hat{y}_{i}^{S} \vert \mathbf{v}^{S}_i \right)
	\end{aligned}
\end{equation}
where $ \hat{y}^{I} $ and $ \hat{y}_{i}^{S} $ are golden intent and slot labels, respectively.

\section{Experiments}
\label{sec:experiments}

\begin{table*}[!t]
	\centering
	\resizebox{0.8\textwidth}{!}{
		\begin{tabular}{l|ccc|ccc}
			\hline
			\multirow{2}{*}{\textbf{Models}} & \multicolumn{3}{c|}{\textbf{CAIS}} & \multicolumn{3}{c}{\textbf{\SmpNluDataset}} \\ \cline{2-7}
			& \textbf{Slot ($ F_1 $)} & \textbf{Intent ($ Acc $)} & \textbf{Overall ($ Acc $)}
			& \textbf{Slot ($ F_1 $)} & \textbf{Intent ($ Acc $)} & \textbf{Overall ($ Acc $)} \\
			\hline
			Slot-Gated~\cite{goo2018slot} &{81.13} &{94.37} &{80.83} &{46.96} &{72.41} &{26.72} \\
			SF-ID Network~\cite{e-etal-2019-novel}  &{84.85} &{94.27} &{82.41} &{41.65} &{74.83} &{25.07} \\
			CM-Net~\cite{liu2019cm}  &{86.16} &{94.56} &{-} &{-} &{-} &{-} \\
			Stack-Propagation~\cite{qin2019stack} &{87.64} &{94.37} &{84.68} &{46.98} &{78.03} &{31.56} \\
			\hline
			Our Model &\textbf{88.61} &\textbf{95.16} &\textbf{86.17} &\textbf{51.10} &\textbf{81.41} &\textbf{34.08} \\
			\hline
		\end{tabular}
	}
	\caption{Main Results on CAIS and \SmpNluDataset.}
	\label{tab:Main resuts}
\end{table*}

\subsection{Experimental Setup}
%
%

To verify the effectiveness of our proposed method, we conduct experiments on two Chinese SLU datasets, including CAIS~\cite{liu2019cm} and \SmpNluDataset.\footnote{\url{http://conference.cipsc.org.cn/smp2019/evaluation.html}}
CAIS dataset includes 7,995 training, 994 validation and 1024 test utterances.
\SmpNluDataset{} dataset consists of 2576 training and 1033 test utterances.

Following~\newcite{Goo}{goo2018slot} and~\newcite{Qin}{qin2019stack}, we evaluate the Chinese SLU performance of slot filling with F1 score and the performance of intent prediction with accuracy, and sentence-level semantic frame parsing with overall accuracy. 
Overall accuracy denotes that the output for an utterance is considered as a correct prediction if and only if the intent and all slots exactly match its ground truth values.

We adopt the Chinese natural language processing system (Language Technology Platform, LTP~\cite{che2020n}), to obtain the word segmentation.\footnote{\url{http://ltp.ai/}}
We use Adam optimizer~\cite{kingma2014adam} with the learning rate of 1e-3 to optimize all trainable parameters.

\begin{table*}[!t]
	\centering
	\resizebox{0.8\textwidth}{!}{
		\begin{tabular}{l|ccc|ccc}
			\hline
			\multirow{2}{*}{\textbf{Models}} & \multicolumn{3}{c|}{\textbf{CAIS}} & \multicolumn{3}{c}{\textbf{\SmpNluDataset{}}} \\ \cline{2-7}
			& \textbf{Slot ($ F_1 $)} & \textbf{Intent ($ Acc $)} & \textbf{Overall ($ Acc $)}
			& \textbf{Slot ($ F_1 $)} & \textbf{Intent ($ Acc $)} & \textbf{Overall ($ Acc $)} \\
			\hline
			w/o Multiple Levels &{82.91} &{95.06} &{80.43} &{50.71} &{77.64} &{32.62} \\
			w/o Sentence-Level word adapter &{86.31} &{94.96} &{83.99} &{50.26} &{79.96} &{33.01} \\
			w/o Character-Level word adapter &{88.28} &{94.66} &{85.18} &{48.96} &{81.22} &{32.14} \\
			\hline
			Full Model &\textbf{88.61} &\textbf{95.16} &\textbf{86.17} &\textbf{51.10} &\textbf{81.41} &\textbf{34.08} \\
			\hline
		\end{tabular}
	}
	\caption{Ablation study on CAIS and \SmpNluDataset{} datasets.}
	\label{tab:Ablation study}
\end{table*}

\subsection{Main Results}
Table~\ref{tab:Main resuts} shows the main results of the proposed model and all compared baselines on the CAIS and \SmpNluDataset{} datasets.
From the results, we have the following observations:
(1) Our model outperforms all baselines on all metrics by a large margin on two datasets and achieves state-of-the-art performance, which demonstrates the effectiveness of our proposed multi-level word adapters;
(2) Our model achieves 0.97\% improvement on Slot ($ F_1 $) score, 0.79\% improvement on Intent ($ Acc $) on CAIS dataset, compared with SOTA model \textit{Stack-Propagation}.
On the \SmpNluDataset{} dataset, we achieve 4.12\% and 3.38\% improvements on Slot ($ F_1 $) score and Intent Accuracy.
These improvements indicate that incorporating word information with proposed multi-level word adapters can benefit the Chinese intent detection and slot filling;
(3) In particular, we obtain the obvious improvements on sentence-level semantic frame accuracy over CAIS (+1.49\%) and \SmpNluDataset{} (+2.52\%) datasets.
We achieve significant improvements due to 1) considering the correlation between two tasks and 2) the mutual enhancement with joint training.

\subsection{Analysis}
%
To verify the effectiveness of the proposed word adapters, we conduct the following ablation analysis.

\newparagraph{Effect of Sentence-Level Word Adapter}
We only remove the \textit{\seqLevel{}} word adapter for intent detection.
The \textit{w/o Sentence-Level word adapter} row in Table~\ref{tab:Ablation study} witnesses an obvious decline for the intent accuracy on ECDT-NLU, which demonstrates that the proposed word adapter can extract word information to promote the Chinese intent detection.

\newparagraph{Effect of Character-Level Word Adapter}
From the results in \textit{w/o Character-Level word adapter} row, we can observe 2.14\% and 0.33\% drops on slot F1 score on ECDT-NLU and CAIS, respectively.
The drops show that word information can help the model to detect the word boundary, and the proposed \textit{\charLevel{}} word adapter can successfully extract word information to improve slot filling.

\newparagraph{Effect of Multiple Levels}
To investigate the effectiveness of the proposed multi-level mechanism, we conduct an experiment to remove the \textit{\charLevel{}} word adapter and provide the same word information for all slots, which is named as \textit{w/o multiple level}. 
From the results shown in Table~\ref{tab:Ablation study}, we can observe that the \textit{\charLevel{}} word adapter bring the significant positive effects.
The speculation that each token needs different word information at a fine-grained level can explain the effectiveness of incorporating token-level word information for \charLevel{} slot filling task.

\section{Related Work}
\label{sec:related_work}

\newparagraph{Intent Detection and Slot Filling}
Different methods, including support vector machine (SVM), recurrent neural networks (RNN), and attention-based models~\cite{haffner2003optimizing,sarikaya2011deep,xia2018zero}, have been proposed to perform intent detection.
For slot filling, conditional random fields (CRF) and RNN were first proposed, and recent approaches are based on self-attention mechanisms~\cite{shen2018disan,tan2018deep}.

Recent works adopted joint models to consider their close relationship between slot filling and intent detection~\cite{zhang2016joint,hakkani2016multi,liu2016attention,goo2018slot,li2018self,qin2019stack}.
However, the above works are restricted to the English language.
In contrast, we consider injecting word information for Chinese SLU in a joint model.

\newparagraph{Chinese Spoken Language Understanding}
To avoid the imperfect CWS and leverage the character features,~\newcite{Liu}{liu2019cm} proposed a character-based joint model to perform Chinese SLU.
Compared with the work, we propose a multi-level word adapter to incorporate word information at both sequence level and character level for intent detection and slot filling.
Word information has shown effectiveness in various Chinese NLP tasks including Chinese word segmentation~\cite{zhang2018chinese,gui2019lexicon} and Chinese text classification~\cite{qiao2019word,tao2019radical}.
These work can be regarded as applications of incorporating word information into subproblems in SLU. 
In contrast with their work, we exploit mutual benefits between intent detection and slot filling for fine-grained word knowledge transfer. To the best of our knowledge, we are the first to incorporate word information to enhance Chinese SLU in a  joint model.
\section{Conclusion}
\label{sec:conclusion}

In this work, we proposed a simple but effective method named multi-level word adapter to inject word information for Chinese SLU tasks.
In particular, we introduced a \textit{\seqLevel{}} word adapter to inject word information for intent detection and a \textit{\charLevel{}} word adapter to incorporate word information for slot filling.
To the best of our knowledge, we are the first to explore injecting word information into Chinese SLU.
The experimental results on two Chinese SLU datasets showed that our model improves performance significantly and achieves the best results.
\section{Acknowledgements}
\label{sec:acknowledgements}

This work was supported by the National Key R\&D Program of China via grant 2020AAA0106501 and the National Natural Science Foundation of China (NSFC) via grant 61976072 and 61772153.
This work was also supported by the Zhejiang Lab's International Talent Fund for Young Professionals.

%


\bibliographystyle{IEEEbib}
\bibliography{refs}

\end{CJK*}
\end{document}